\documentclass[10pt,twocolumn,letterpaper]{article}

\usepackage{wacv}
\usepackage{times}
\usepackage{epsfig}
\usepackage{graphicx}
\usepackage{amsmath}
\usepackage{amssymb}

\wacvfinalcopy 

\def\wacvPaperID{433} 

\ifwacvfinal\fi
\begin{document}

\title{Generalization in Metric Learning: \\ Should the Embedding Layer be Embedding Layer?}

\author{Nam Vo \\
Georgia Tech \\
{\tt\small namvo@gatech.edu}
\and
James Hays \\
Georgia Tech \\
{\tt\small hays@gatech.edu}
}

\maketitle
\ifwacvfinal\thispagestyle{empty}\fi

\begin{abstract}


This work studies deep metric learning under small to medium scale data as we believe that better generalization could be a contributing factor to the improvement of previous fine-grained image retrieval methods; it should be considered when designing future techniques. In particular, we investigate using other layers in a deep metric learning system (besides the embedding layer) for feature extraction and analyze how well they perform on training data and generalize to testing data. 
From this study, we suggest a new regularization practice where one can add or choose a more optimal layer for feature extraction. State-of-the-art performance is demonstrated on 3 fine-grained image retrieval benchmarks: Cars-196, CUB-200-2011, and Stanford Online Product.

\end{abstract}


\section{Introduction}

We study small to medium scale deep metric learning (DML) with application to fine-grained image retrieval, from the perspective of generalization. In particular, analyzing training performance could lead to new insight; for example, He at al \cite{he2016deep} showed deeper network under-perform on training data and proposed residual connection to overcome that obstacle.

Deep learning has helped to advance many computer vision tasks, including fine-grained image retrieval: identifying reference images semantically similar to the input/query image. The state of the art approach is to learn an image feature extractor network by using a DML loss function that encourages semantically similar images to have similar features. Hence retrieval can be efficiently done by looking up the nearest neighbors in the feature space.

\begin{figure}
  \includegraphics[width=250pt]{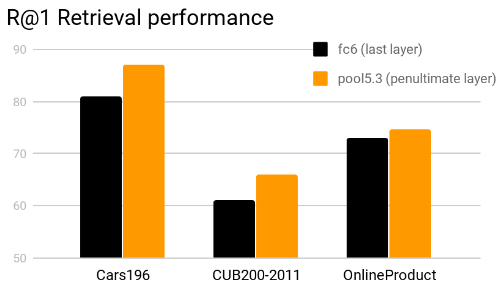}
  \caption{Recall at rank 1 performance on 3 benchmarks when using the last layer (fc6) vs the penultimate layer (pool5.3)}
  \label{fig:intro}
\end{figure}

Given a trained network (for any task, not necessary DML only), the output of any layer can be used as image feature. DML works often use the ``embedding'' layer: the last one in the network that is the input of the loss function. We train a network using DML and decide to measure the performance of other layers; interestingly the layer before the embedding layer often outperform it, as shown in Figure \ref{fig:intro}. Upon further investigation, it is found that while the last layer performs the best on training data, it doesn't generalize well.

Motivated by above observation, we present here a study of generalization in the context of small to medium scale DML. Niche areas such as fine-grained image retrieval might not have the usual data abundance assumption, so common practice is to finetune a network that is pretrained on another large-scale dataset. In such scenario, it is very easy to fit the training data and so the system's generalization ability is an important factor.

We explore the following hypotheses:

1. The embedding layer is not pretrained: so it does not generalize as well as the penultimate layer (pool5.3).

2. Pool5.3 layer is shallower: shallow networks fit training data worse but generalize better than deeper networks.

3. Distance from the loss layer: the closer a layer to the loss, the better its feature fits training data and the worse it generalize to test data.

We show, from our analysis, how to train a simple DML pipeline that outperforms state-of-the-art on 3 fine-grained image retrieval benchmarks CUB-200-2011, Cars-196, and Stanford Online Products. Note that we are not proposing a new method; the finding here is potentially applicable to previous works to improve their performance, as shown in our experiments.

\section{Related Works}

Deep metric learning in computer vision has many applications, from generic image search \cite{wang2014learning} to sketch-based search \cite{sangkloy2016sketchy}, image geolocalization \cite{vo2016localizing, vo2017revisiting}, people re-identification \cite{ding2015deep,hermans2017defense} and face recognition \cite{schroff2015facenet, Parkhi15}.

The popular approach is to use a Siamese \cite{bromley1994signature,chopra2005learning} or triplet network \cite{wu2013online,wang2014learning,vo2016localizing}: training examples are put in the form of pairs or triplet, with the label being whether they are similar or not; and either the contrastive loss or triplet hinge loss is used for training, it encourages similar images to have similar features and dissimilar images to have different features.
Most recent advances of DML focus on example mining strategy and/or better loss function.

The Multibatch method \cite{tadmor2016learning} performs metric learning on all possible pairs from the mini-batch and show that it reduces the variance of the estimator and significantly speed up the convergence rate. In \cite{vo2016localizing},
a similar trick called mini-batch exhausting improves retrieval performance by simply extending the learning to all possible triplets in the mini batch. In \cite{hermans2017defense}, it is studied under the name of batch-all and batch-hard. Kihyuk Sohn \cite{sohn2016improved} used a similar batch construction strategy and proposed the multiclass n-pair loss which improves upon the triplet loss.

In \cite{kumar2017smart}, Harwood et al proposed a smart mining procedure to select effective samples from the whole training data; this helps the training to converge faster. 
Huang el al \cite{huang2016local} proposed a position-dependent deep metric unit to adaptively select hard examples taking into account the local neighborhood.
In \cite{wu2017sampling}, Wu et al propose to sample uniformly w.r.t. 
their relative distance; this distance weighted sampling is shown to be more effective than random or the common semi-hard sampling.

Wang et al \cite{wang2017deep} proposed a new loss that constraints the angle of the triplet triangle and demonstrate favorable properties over the traditional distance-based hinge loss for triplet. Different from previous works, the proxy-based loss \cite{movshovitz2017no} behaves like a classification loss; each training example is compared against learned proxies, not with each other in a pair, triplet or cluster, hence eliminate the need of example mining.

In \cite{opitz2017bier}, multiple embeddings are learned at the same time, later one would focus more on examples that are deemed hard to previous ones; the output embedding is a concatenation of all learned embedding.
In a similar spirit, Yuan et al \cite{yuan2016hard} propose to learn a set of embeddings organized in a cascaded manner, in which easy training examples are filtered out sequentially and only the hardest one reach the last embedding loss. 

\section{Studying How Well a Layer Generalizes}

\begin{figure*}
  \includegraphics[width=\textwidth]{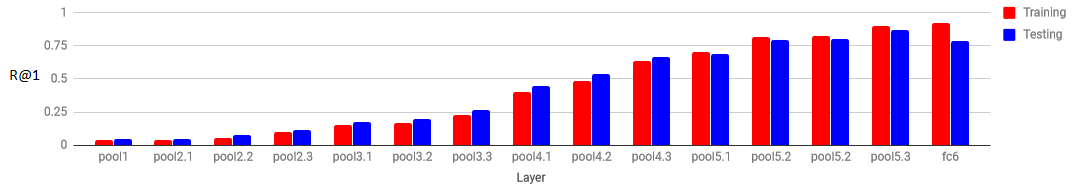}
  \caption{R@1 performance of different layers on training and test set of Cars-196.}
  \label{fig:layers}
\end{figure*}

\begin{figure}
  \includegraphics[width=240pt]{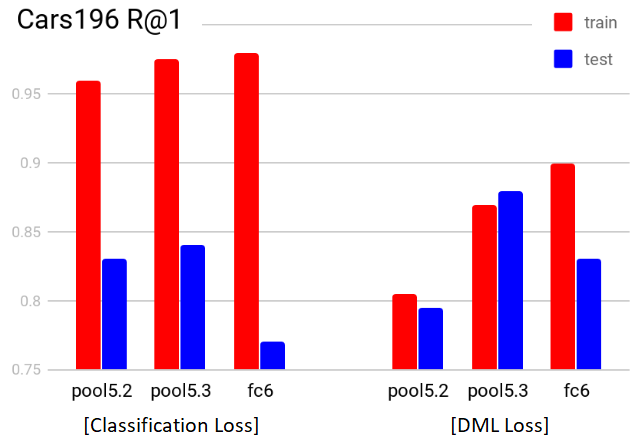}
  \caption{R@1 performance of different layers on training and test set of Cars-196. Left: the variant trained with classification loss. Right: our baseline network trained with DML loss.}
  \label{fig:classification}
\end{figure}

We'll start with the experiments result and analysis on the small scale dataset Cars196 \cite{KrauseStarkDengFei-Fei_3DRR2013}. Further experiments on other datasets will be discussed later.

The set up will be the image retrieval task; we want to learn a deep network for extracting image features such that images that are semantically similar are close to each other.
The common metric Recall at rank 1 (R@1) is used, which is the percentage of test images whose first nearest neighbor (in the feature space) is of the same class.

Applying DML approach, we train the VGG network from an ImageNet-pretrained model. We will describe our implementation in the next section, but note that we are not proposing a new method, the observation here is consistent with different DML systems. 

In Figure \ref{fig:layers}, we show the retrieval performance on training data and testing data when using output of different layers as the image feature (we perform max pooling so that all features have similar dimensionality of 512, the raw feature outputs of early layers are too big to be practical and actually perform worse). We observe the performance increases as deeper layers in the network are used, except the last layer doesn't improve upon the layer before it on test data. In Figure \ref{fig:classification}-left, we plot the performance of the last 3 layers (pool5.2, pool5.3, and fc6) changing as the network is trained: although fc6 improves and surpass pool5.3 on training data as expected, it doesn't generalize as well, eventually resulting in worse test time performance. This goes back to the fact that small to medium scale training is more susceptible to over-fitting than under-fitting.

\subsection{The role of better loss and training strategy}

Our system has no problem over-fitting such small-scale training data if trained long enough (for example over 90\% R@1 can be achieved in 100 epochs). We speculate that it is similarly easy for other systems to overfit. So a contributing factor to previous improvements could be better generalization, which can be affected by the loss function, training and architecture design or regularization techniques.

We provide an example: we train the same baseline network, but replace the DML loss with classification-output-fc7 and classification-loss (Softmax-CrossEntropy); the result is shown in Figure \ref{fig:classification}-right. Under this loss, the fitting and generalizing behavior is quite different: all 3 layers fit better, but generalization is worse.

Note that classification loss has been previously shown to help improving retrieval performance in multi-task/co-training, especially in large-scale training setting where fitting data really well is the more important factor \cite{Parkhi15, wen2016discriminative}. Even though the experiment result is different from ours, Horiguchi et al \cite{horiguchi2017significance} argue that classification loss is advantageous when the size of training data (samples per class) is large. In \cite{vo2017revisiting}, it is argued that learning with classification is more effective than metric learning because training data is very diverse/noisy.


\subsection{Is pretrained layer better for embedding}

\begin{figure*}[h!]
\begin{center}
  \includegraphics[width=\textwidth]{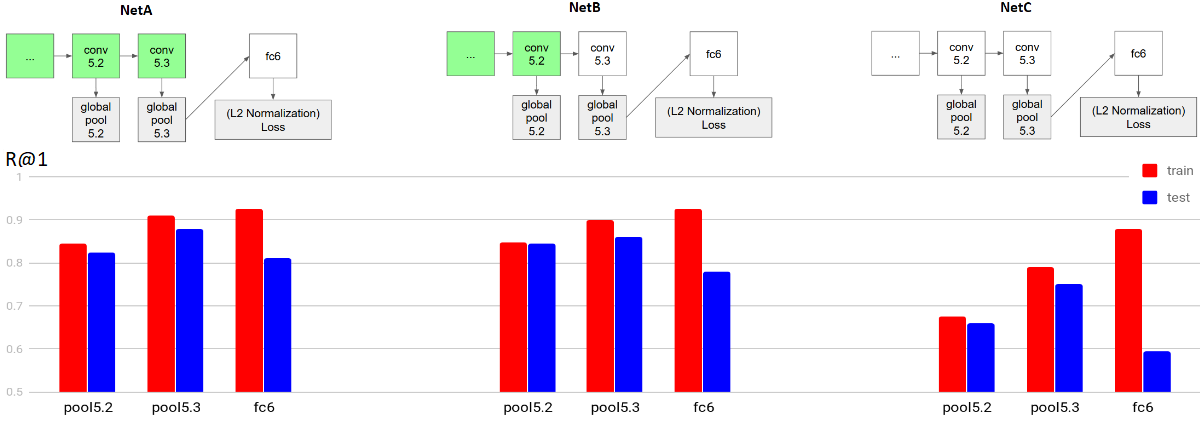}
  \vspace{5pt}
  \caption{R@1 performance of different layers on training and test set of Cars-196. Green box: pretrained layer, white box: initialized from scratch layer, gray box: parameterless layer. Best viewed in color.}
  \label{fig:analysize1}
\end{center}

\begin{center}
  \includegraphics[width=\textwidth]{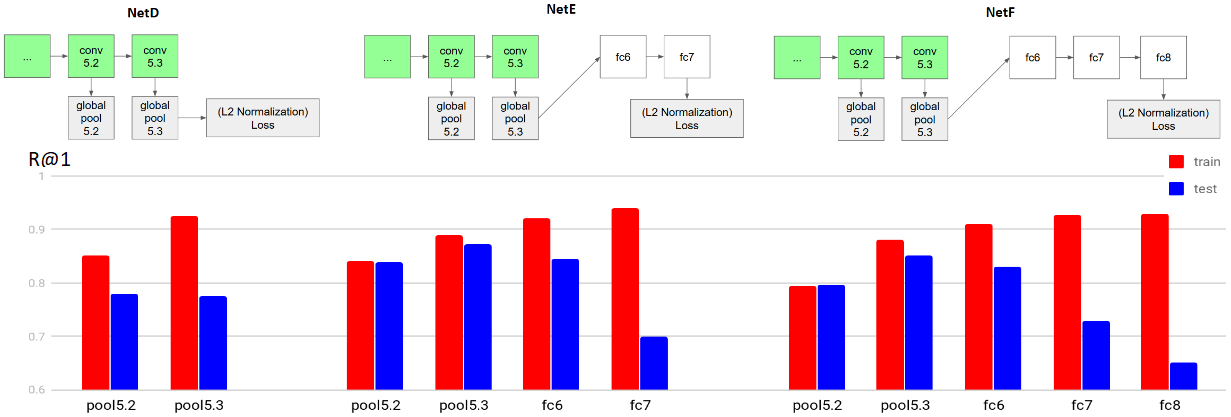}
  \vspace{5pt}
  \caption{R@1 performance of different layers on training and test set of Cars-196. Compared to NetA in Figure 4, the position of the loss function is changed. Best viewed in color.}
  \label{fig:analysize2}
\end{center}
\end{figure*}

Now we know that pool5.3 can generalize better than fc6; the first possible explanation for that is: conv5.3 (and all layers before it) is pretrained on ImageNet while the embedding layer fc6 is added later for this task and initialized from scratch. We compare 2 cases in Figure \ref{fig:analysize1}: NetA, the same baseline we are using, and NetB, where both conv5.3 and fc6 are initialized from scratch; for reference, we also include the case of NetC where the whole network is initialized from scratch.
It can be seen that with one or more layers initialized from scratch, the performance of most degrades accordingly. Except for fc6, directly under loss layer, can continue fit the training data better if trained long enough.

The training performance of fc6 is better than pool5.3 which in turn is better than pool 5.2. This is expected, as the feature obtained from a layer that is closer to the loss layer work better on training data.

Performance on test data is more interesting:
\begin{itemize}
  \item NetB:pool5.2 is much better than NetA:pool5.2. It could be conv5.3 being initialized from scratch has a regularization effect. So pretraining a layer might be good for layers after it but has a "bad effect" for layer before it?
  \item Pool5.3 performs the best, whether it is pretrained or not. In fact NetB:pool5.3 outperforms both NetB:fc6 and NetA:fc6. It is also better than the NetB:pool5.2 (where conv5.2 is pretrained).
\end{itemize}

So while conv5.3 being pretrained does indeed help improve performance, this alone does not explain why pool5.3 is better than fc6.

\subsection{The depth and the loss layer's position}

We explore other potential explanation: layer's depth and the position of loss function. Speculating that the closest layer to loss function might be more vulnerable to overfitting, we move the loss layer up/down and introduce 3 new cases, shown in Figure \ref{fig:analysize2}.
\begin{itemize}
  \item NetD: similar to NetA except we remove fc6, so the loss is now directly after pool5.3.
  \item NetE: similar to NetA except we add another fully connected layer fc7.
  \item NetF: similar to NetA except we add 2 fully connected layer fc7 and fc8.
\end{itemize}

All new layers have the same output size 512 with ReLU activation between them (not shown in the figure).

First, adding more new layers seems to worsen test time performance; hence the common practice of adding 1 new layer only when finetuning a pretrained model for another task.

A layer performs worse when it is the last layer than when it is the penultimate: pool5.3 in NetD is significantly worse than NetA (even though it's slightly better on training data). Similarly for fc6 (NetA vs NetE) and fc7 (NetE vs NetF) although the difference is smaller.

The last observation is that: when being the penultimate layer and initialized from scratch, pool5.3 (NetB) and fc6 (NetE) performance are somewhat similar. When being the last layer, pool5.3 (NetD) can be slightly worse than fc6 (NetA).

\subsection{Switching or adding layer as a regularization}


The position of the loss layer w.r.t. the embedding layer seems to play an important role in explaining why one layer is better than the other. Hence to get the best possible test time performance, one should consider cross validation to choose the best layer. If the output layer is fixed, consider making it the penultimate, so that the loss layer has a less direct influence, therefore reducing overfitting.

As a good practice, we suggest using NetA:fc6 and NetA:pool5.3 respectively in the cases of less over-fitting to more over-fitting. In small-scale setting, NetE:pool5.3 can also do well; and NetE:fc6 can be used when one can not control the pretraining step but want to define the embedding dimension. Note that when considering these options, it should be taken into account that their behavior can be different with not only different dataset/task, but also different loss function and training strategy.

\section{Implementation Detail}

Our thesis is that generalization is an important factor. With that in mind, we describe the DML design that we choose to work with. Note that we are not proposing a new method, all components here already exist or can be easily implemented; though we will release the source code and models from this study. 

\subsection{Base architecture}

We use the VGG-16 architecture with conv layers only (fc6, fc7 and fc8 are removed), followed by a global pooling layer. This architecture is popular for scene image retrieval task \cite{arandjelovic2016netvlad,tolias2015particular,radenovic2016cnn,gordo2016deep}. Although more sophisticated pooling method can be better (NetVLAD \cite{arandjelovic2016netvlad}, R-MAC \cite{tolias2015particular} or GeM \cite{radenovic2017fine}), we use the simple MAX pooling.

We employ BatchNorm \cite{ioffe2015batch} in our system, which is known to make training efficient and also provide regularization effect. Greff  et al \cite{greff2016highway} shows, with ResNet and Highway network making optimization easier, the role of BatchNorm becomes purely regularization (higher training loss, lower testing error). Dropout is also experimented with but not as effective as a regularizer.

We use PyTorch as the framework for experimenting, which comes with VGG-BatchNorm pretrained on ImageNet.

\subsection{Feature extraction layers}

A new fully connected layer fc6 is added at the end of the network, this is the traditional output layer or embedding layer. In this work we experiment with using, not only this last layer, but also other layers before it as the feature extraction layer.

In the experiment from previous section, we also have other variants (NetE and NetF) where we further add fc7 and fc8. All of them have the same output size of 512.

\subsection{Training mini-batch construction}

While DML loss often involves looking at pair or triplet as the example, recent works construct a batch of images as input instead, and only form pairs/triplets/clusters at the loss layer. Hence the mini-batch can be constructed randomly in any way as long as a lot of similar/dissimilar pairs/triplet can be formed (for example \cite{vo2016localizing, sohn2016improved}).

In case the data comes with the class label, we can randomly sample p classes ($p>1$), k images per class ($k>1$) resulting in a batch size of $m=pk$. Under fixed m, maximizing $p$ will increase the diversity leading to more stable gradient; even though the number of possible triplets, $pk(k-1)(pk-k)$, is reduced.

\subsection{Loss function and example mining}
Aside from \cite{radenovic2016cnn} and \cite{kumar2017smart} that actually perform hard example mining on the whole training data, most mining strategies operate within mini-batch, which is constructed randomly uniform. Hence all these approaches can simply be formulated as a loss function operating at mini-batch level looking at all possible pairs/triplets/clusters and weight them differently. For instance, the recently proposed focal loss for classification task \cite{lin2017focal} quickly diminish the contribution of easy examples.

Hence we are not doing any explicit hard example mining.
Given the batch $b$ of output image features, our loss is defined as:

$$L(b) = \frac{1}{M} \sum_{triplet (a,p,n) \subset b}^{} tripletloss(a, p, n)$$

where $M = \sum_{triplet (a,p,n) \subset b}^{} 1$, the total number of valid triplets in a batch; (a, p, n) is a triplet of the anchor, the similar and the dissimilar instance respectively. We follow \cite{vo2016localizing, sohn2016improved, hermans2017defense} and use the smooth loss function (which shown to be better than the traditional hinge loss):

$$tripletloss(a, p, n) = log(1 + exp(d(a,n) - d(a,p)))$$

Where d is the squared Euclidean distance (negative dot product can also be used instead). Note that it is sensitive to the scale of the image feature. In our implementation, we normalize the image feature to have unit magnitude and then scale it by 4, as suggested by \cite{vo2016localizing}. An alternative is to not do normalization and let the network learn the scaling, as suggested \cite{hermans2017defense}; which also works in our experience, though one has to be careful with the initialization to avoid numerical instability at the beginning.

\section{Experiments}

\begin{figure}
  \includegraphics[width=220pt]{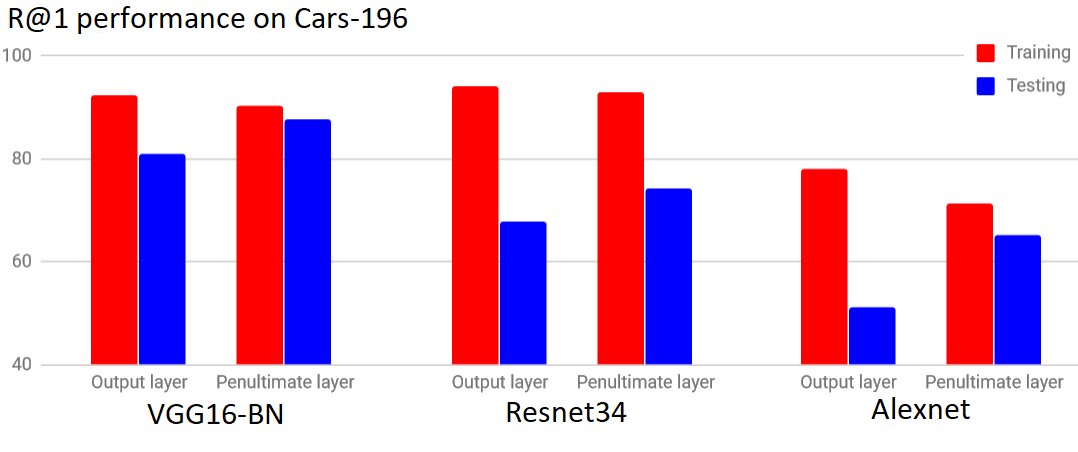}
  \vspace{2pt}
  \caption{Comparing the performance of output layer vs the layer before it when using 3 different backbone architectures.}
  \label{fig:architectures}
\end{figure}

\begin{figure}
  \includegraphics[width=240pt]{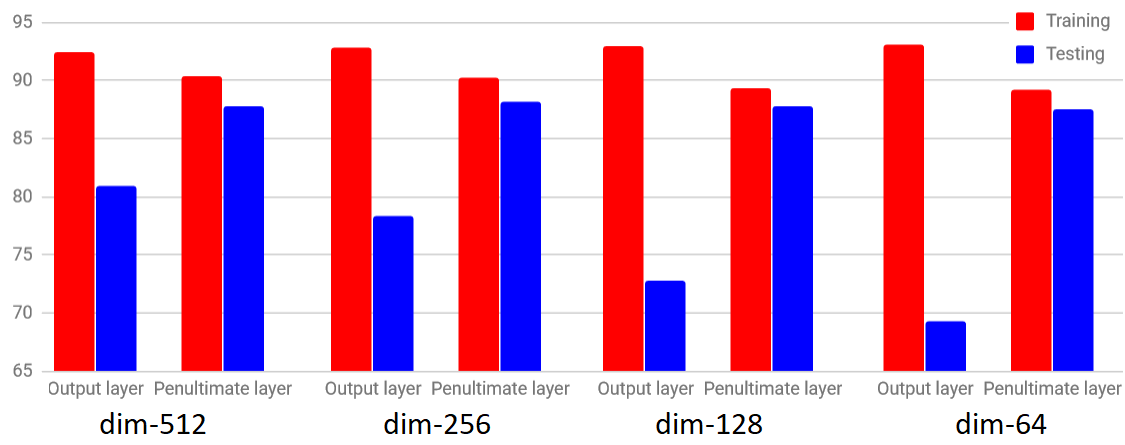}
  \vspace{2pt}
  \caption{Comparing the performance of output layer vs the layer when varying the output layer's dimensionality.}
  \label{fig:dims}
\end{figure}

\begin{figure*}[h]
\includegraphics[width=0.99\linewidth]{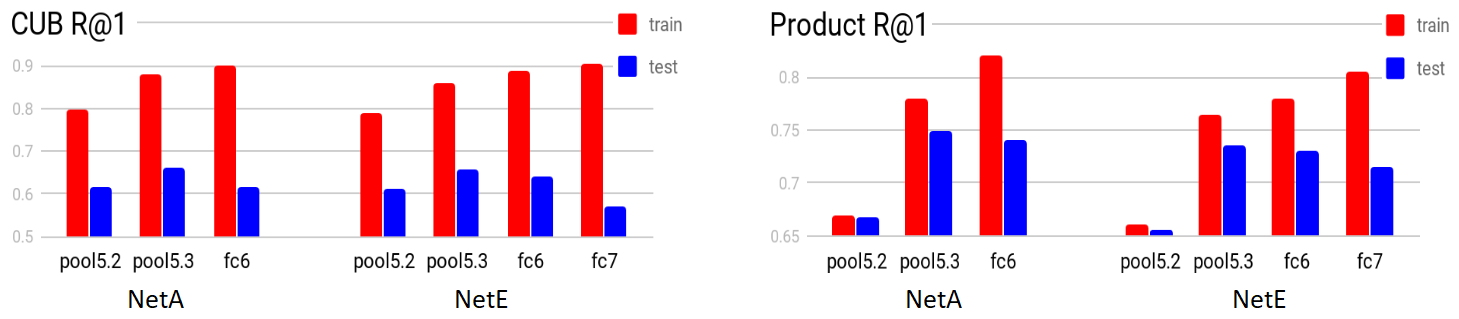}
\caption{R@1 performance of NetA and NetE on CUB-200-2011 (left) and Stanford Online Product (right)}
\label{fig:cub}
\end{figure*}

\begin{figure*}
\begin{center}
  \includegraphics[width=0.85\linewidth]{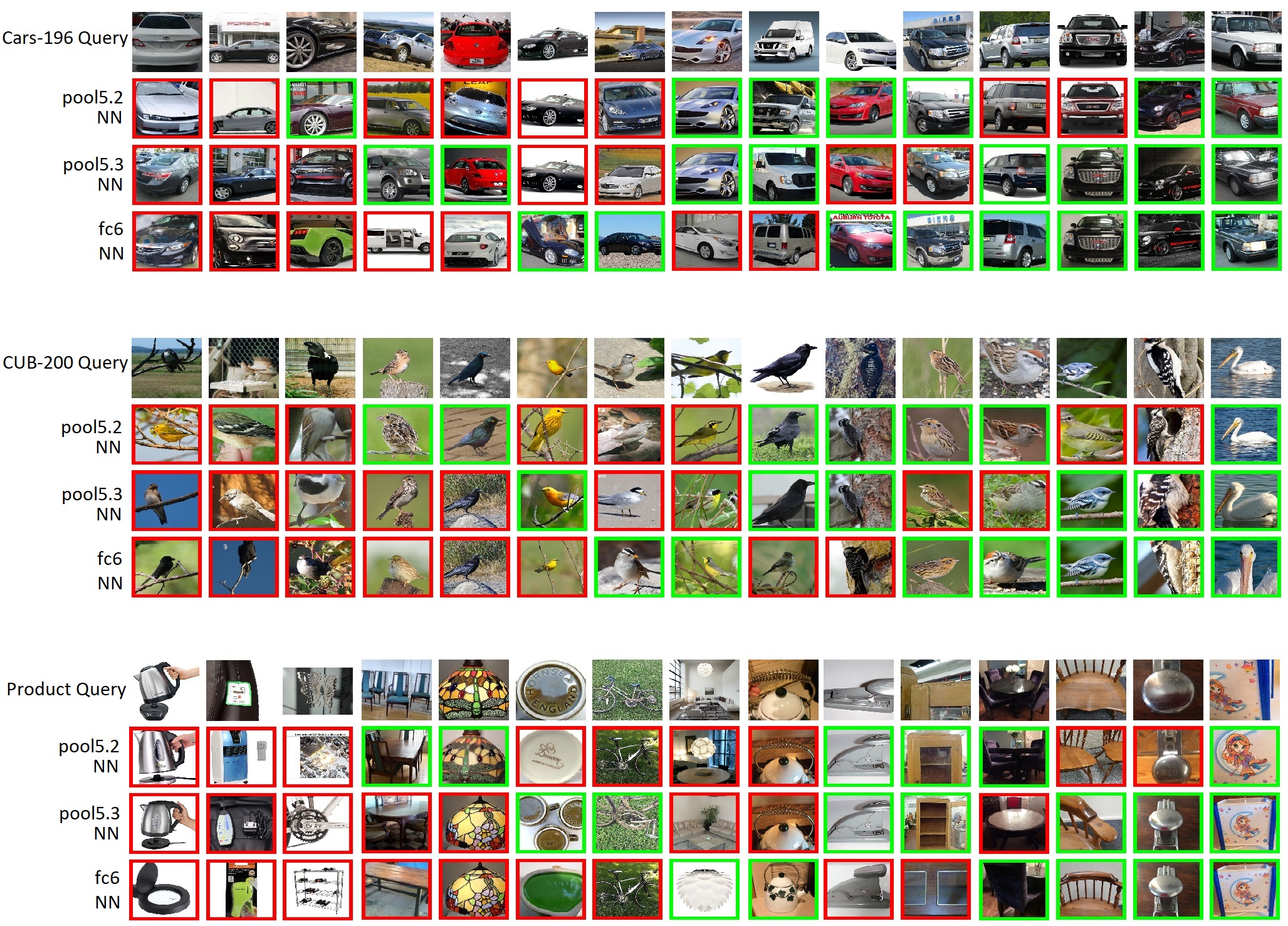}
  \caption{Some nearest neighbor (NN) retrieval examples on the 3 datasets, we show cases in which using feature from different layers results in different NN.}
  \label{fig:qual}
\end{center}
\end{figure*}

\setlength{\tabcolsep}{5.1pt}
\begin{center}
\begin{table*}[]
\begin{center}
\centering
\label{table:benchmarks0}
\caption{R@k performance on 3 benchmarks.}
\begin{tabular}{| l || cccccc || c ccccc || c ccc |}
\hline
Dataset  & \multicolumn{6}{c||} {Cars-196} 
       & \multicolumn{6}{c||} {CUB-200-2011} 
       & \multicolumn{4}{c|} {Product} 
       \\
\hline

k  &    1 & 2 & 4 & 8 & 16 & 32
        &  1 & 2 & 4 & 8 & 16 & 32
        &  1 & 10 & 100 & 1000
       \\
       
\hline
\hline

NetA:pool5.3  &  87.8 & 92.7 & 95.6 & 97.5 & 98.6 & 99.2
          &  66.4 & 77.5 & 85.4 & 91.3 & 95.2 & 97.1
          &  74.8 & 88.3 & 95.2 & 98.5 \\

\hline

\end{tabular}
\label{table:benchmarks0}
\end{center}
\end{table*}
\end{center}

\setlength{\tabcolsep}{5.1pt}
\begin{center}
\begin{table*}[]
\begin{center}
\label{table:benchmarks1}
\centering
\caption{R@1 performance on 3 benchmarks, compared to previous works. }

 \vspace{5pt}

\begin{tabular}{| l | ll | ccc |}
\hline
Method  & Network & dim & Cars-196 & CUB-200 & Product   \\

\hline
\hline
Lifted structure \cite{oh2016deep}  & GoogLeNet & 64/64/512  & 53.0  & 47.2 & 62.5  \\
Facility \cite{songCVPR17}  & Inceptionv1BN & 64 &  58.1 & 48.2 & 67.0  \\
Angular loss \cite{wang2017deep}  & GoogLeNet & 512 &  71.4 & 54.7 & 70.9  \\
HDC \cite{yuan2016hard}  & GoogLeNet & 128x3 &  73.7 & 53.6 & 69.5  \\
BIER \cite{opitz2017bier}  & GoogLeNet & 512 &  78.0 & 55.3 & 72.7  \\
Margin \cite{wu2017sampling}  & ResNet50 & 128 &  \underline{79.6} & \textbf{63.6} & 72.7  \\
Proxy-NCA \cite{movshovitz2017no}  & InceptionBN & 64 &  73.2 & 49.2 & \underline{73.7}  \\
  
ABE \cite{kim2018attention}  &  8 Heads Ensemble & 512 &  \textbf{85.2} & \underline{60.6} & \textbf{76.3}  \\

\hline

& & & & & 
 \vspace{-5pt}
 
\\

Our NetA     & VGG16-BN (layer pool5.3) & 512 &  \textbf{87.8} & \textbf{66.4} & \textbf{74.8}  \\
 & VGG16-BN (layer fc6) & 512 &  81.0 & 61.7 & \underline{74.1}   \\
     
 Our NetE    & VGG16-BN (layer pool5.3) & 512 &  \underline{86.7} & \underline{65.8} & 73.6  \\ 
     & VGG16-BN (layer fc6) & 512 &  84.3 & 63.7 & 73.3   \\
  & VGG16-BN (layer fc7) & 512 &  71.4 & 56.8 & 71.4   \\

\hline

\end{tabular}

\label{table:benchmarks1}
\end{center}
\end{table*}
\end{center}

Here we provide more detail about the experiment set up on 3 fine-grained image retrieval datasets:
\textbf{Cars196} \cite{KrauseStarkDengFei-Fei_3DRR2013}: consist of 16k images of 196 different classes of cars. Following the typical protocol, we train on the first 98 classes and test on the other 98 classes.
\textbf{CUB-200-2011} \cite{WahCUB_200_2011}: has around 12k images of 200 bird species. The first 100 classes (5864 images) are used for training and the rest for testing.
\textbf{Stanford Online Product} \cite{song2016deep}: is bigger than the other 2 with 120k images of more than 22k products from eBay. Similarly, the first half (11318 classes with a total of 59551 images) are used for training and the remaining classes are for testing.

We use stochastic gradient descent with momentum weight 0.9, weight decay factor 5e-4, learning rate 0.01 (decreasing by 10 times during training). Random scaling, cropping, and flipping are used for data augmentation; a single center crop is used during test time. We use batch size 32, train for 20k iterations in case of Cars-196 and CUB-200-2011, and 200k iterations in case of Online Product.

\subsection{Ablation study}

We switch the backbone architecture from VGG-16-batchnorm to Alexnet and Resnet34. As shown in Figure \ref{fig:architectures}, the observation is still similar: the penultimate layer is better than output layer during test time; though the performance is lower compared to using VGG.

We lower the output dimensionality from 512 to 256, 128 and 64. As shown in Figure \ref{fig:dims}, this degrades the output layer fc6 performance, while the penultimate layer pool5.3 (still with dimensionality 512) is affected slightly.

The soft triplet based loss we used are popular and work well across different image retrieval tasks. In our experience, the contrastive loss, aslo popular, is slightly less effective. The classification based loss fits extremely well, but can be more vulnerable to overfitting, as shown in figure \ref{fig:classification}.

\subsection{Result}

In Figure \ref{fig:cub}-left, we show NetA and NetE performance on CUB-200-2011. This dataset is much harder to generalize even though its size is similar to that of Cars-196. Still, pool5.3 is the best performing layer, and fc6 does better if it is not next to the lost layer.

In Figure \ref{fig:cub}-right, we show performance on Stanford Online Product. Similar to previous cases, fc6 is outperformed by pool5.3; however due to this dataset's bigger size, there are some differences: (1) it takes much longer to train, (2) NetA seems to fit the training data slightly better/faster than NetE, resulting in better test time performance too. We speculate that when training at an even larger scale, bad generalization will be a less significant concern. We show some retrieval qualitative result in the supplemental.

\textbf{Comparing to state-of-the-arts:} in Table \ref{table:benchmarks1}, we report the R@1 performance of different layers of NetA and NetE. As a baseline, our fc6 yields comparable result to some state of the art approaches, and our pool5.3 result outperform all of them. Note that each component (loss function, mining strategy, etc) of each system might not be directly comparable because of difference in network architecture and embedding dimension (and possibly data preprocessing, training hyper-parameters, etc). Moreover, we are investigating the generalization effect of different layers, not proposing a new method to replace existing works.


\subsection{Analyzing publicly released models}

This effect can also be demonstrated on other models, not just ours. Among previous works, LiftedStructure \cite{oh2016deep} and HDC \cite{yuan2016hard} have released the models used in their papers. We obtained and tested these models, the result is shown in Table \ref{table:lifted}. Similar to our system, the output layers here are also outclassed by the layer before it during test time.

One could argue that, different from our case, it is mainly because of the bigger embedding size here. However notice that output layers still fit quite well on training data even with smaller embedding size, but generalize worse on testing data. \cite{oh2016deep} made an observation that the embedding size did not play a significant role in their system. But we do observe that the size is a contributing factor affecting both fitting and generalizing performance, and not only that of the embedding layer but also other layers' as well.

Finally, we observe that our system generalizes better: at similar training performance, our output layer still perform better on the test set (refer to Figure \ref{fig:analysize1} and \ref{fig:cub}); this is also the case for penultimate layers. We speculate the reason to be the difference in embedding size, loss function, network architecture, and its corresponding pretrained model. Worse generalization might be what prevented previous works from fitting the training data further.

\setlength{\tabcolsep}{5.1pt}
\begin{center}
\begin{table}[]
\begin{center}
\centering
\label{table:lifted}
\caption{R@1 performance of different layers from publicly released lifted structure model. Penultimate denotes the layer before the output layer.}

\vspace{5pt}

\begin{tabular}{| lr | cc | cc | cc |}
\hline

R@1 &  & \multicolumn{2}{c|} {Cars-196}  &  \multicolumn{2}{c|} {CUB-200} &  \multicolumn{2}{c |} {Product}  \\
Layer  &  & train & test & train & test & train & test   \\

\hline
\hline

\multicolumn{8}{| l |} {Lifted Structure released model \cite{oh2016deep}} \\
\hline
model-512  &  & \textbf{97.7} & 47.9  & {66.4} & 46.3 &  65.2 & 62.1  \\
penultimate  &  &  97.0 & \textbf{65.1}  & 66.4 & \textbf{47.1} & \textbf{67.7} & \textbf{65.9}  \\                             

\hline
\hline

\multicolumn{8}{| l |} {HDC released model \cite{yuan2016hard}} \\  
\hline
\multicolumn{1}{|l} {outputs} & & \textbf{80.4}  & 73.7  & \textbf{83.7} & 53.6 & \textbf{76.5} & {69.5}   \\
\multicolumn{1}{|l} {penultimates} &   & 75.5  & \textbf{75.7}  & 79.9 & \textbf{57.2} & 75.6 & \textbf{70.1}  \\
                                                                              
\hline
\hline

\multicolumn{8}{| l |} {Our NetA} \\
\hline
output (fc6)  &  &  \textbf{92.4}  & 81.0  & \textbf{89.6} & 61.7 & \textbf{81.2} & 74.1  \\       
penultimate  &  & 90.4 & \textbf{87.8}  & 87.6 & \textbf{66.4} & 77.9  & \textbf{74.8}  \\                      

\hline

\hline

\end{tabular}
\label{table:lifted}
\end{center}
\end{table}
\end{center}

\subsection{Experiment on revisited Oxford-5k and Paris-6k benchmarks}

We perform additional experiments on 2 popular scene image retrieval datasets: Oxford \cite{Philbin07} and Paris \cite{Philbin08}, using the new labels and benchmarking protocol introduced in \cite{Radenovic-CVPR18}.

Implementation detail: we use the NetA same as previous experiments, but with VGG-16 (without batch-norm) as the backbone architecture. We use the same training data as in \cite{gordo2016deep}, which we manage to collect around 20k images. During training, images are resized and cropped to 500x500.

\begin{figure}
  \includegraphics[width=250pt]{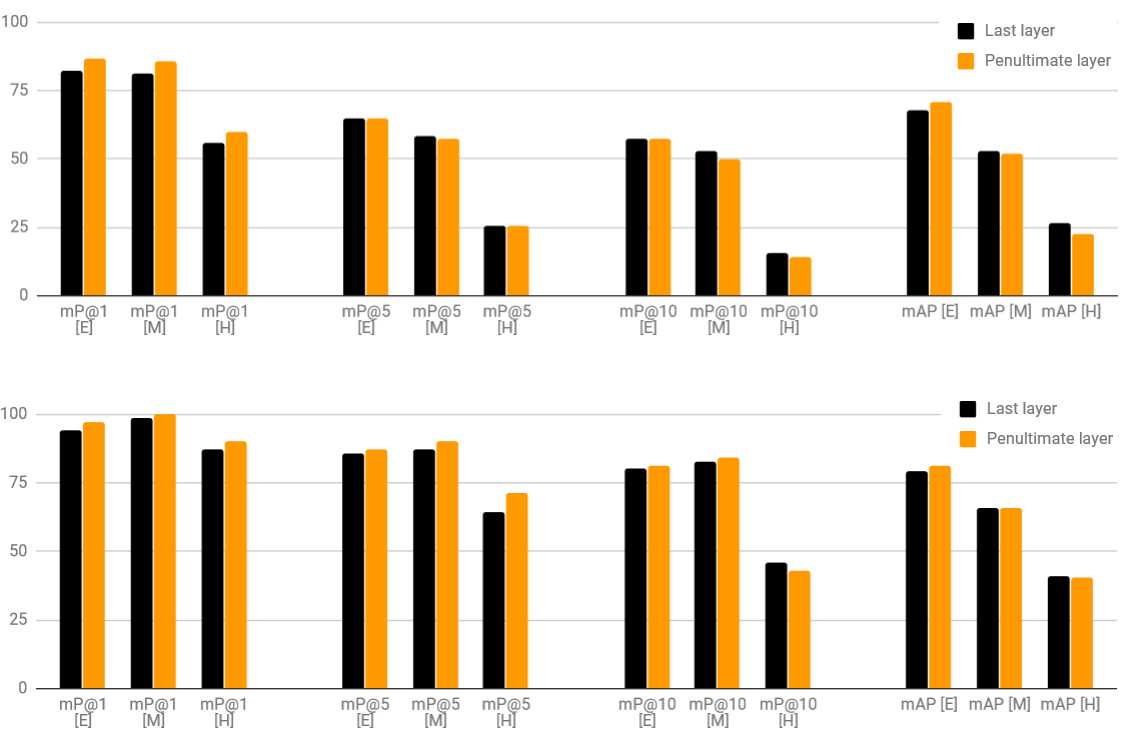}
  \vspace{2pt}
  \caption{Image retrieval performance on ROxford-5k (top) and RParis-6k (bottom), when using last layer vs penultimate layer for feature extraction}
  \label{fig:revi}
\end{figure}

In \cite{Radenovic-CVPR18}, 3 different evaluation setups are introduced: Easy (E), Medium (M) and Hard (H), with several metrics: mP@1, mP@5, mP@10 and mAP (we refer readers to \cite{Radenovic-CVPR18} for more detail). We show the result under these different setups in Figure \ref{fig:revi} when using the last layer vs the penultimate layer.

It can be observed that using the last layer tends to get a worse result in Easy setup or using mP@1 metric, while having the advantage in Hard setup and using mP@10 or mAP metric. Note that mP@1 metric here is the same as the R@1 metric used in the 3 previous fine-grained image retrieval benchmarks, so in this case, it's consistent with previous observation that penultimate layer is better.

However under mAP metric, and especially in Hard mode, using the last layer is better. The explanation could be how differently these metrics reflect the result quality. When the application is to search for relevant reference images (for example "collect all Eiffel tower images in the database") mAP in Hard mode should be used; on the other hand, if the task is using image retrieval to make prediction about the query image (for example "identify this tower" or "recognizing this face"), R@k, R@1 or mP@1 in Easy mode is more appropriate. Our speculation is that, when using penultimate layer, examples of the same class form more complex manifold in this latent space, such that it's easier to search for few neighbors of the same classes, but more difficult to identify all instances of that class.

State of the art systems, with better training data collection procedure, additional bells and whistles, have achieved more impressive results on these benchmarks, we refer readers to \cite{Radenovic-CVPR18} for detail. 

\section{Conclusion}

In this study, we analyze generalization as a big contributing factor to improvements made on small scale fine-grained image retrieval benchmarks. As we have shown, choosing a different loss or different feature extraction layer can improve test time performance even though it actually under-perform on training data, and vice versa. Hence generalization should be kept in mind when designing or analyzing new techniques in the future.

Given that any layer in a network can be used for feature extraction, the major part of this work is an ablation study of how each layer performs differently in the fine-grained image retrieval task. Not too surprising, we found that the closer a layer to the loss layer, the better its produced feature is. Different from what people would assume, the last layer is usually however not the best one at test time, due to bad generalization; the second last layer is. The observation holds across different benchmarks and not only on our model, but also other models released by previous works.

{\small
\bibliographystyle{ieee}
\bibliography{egbib}
}

\newpage
\clearpage
\newpage
\clearpage
\setcounter{equation}{0}
\setcounter{section}{0}
\setcounter{page}{1}

\begin{center}
\textbf{\large Supplemental Material}
\end{center}

\section{Training plots}

\begin{figure}
  \includegraphics[width=240pt]{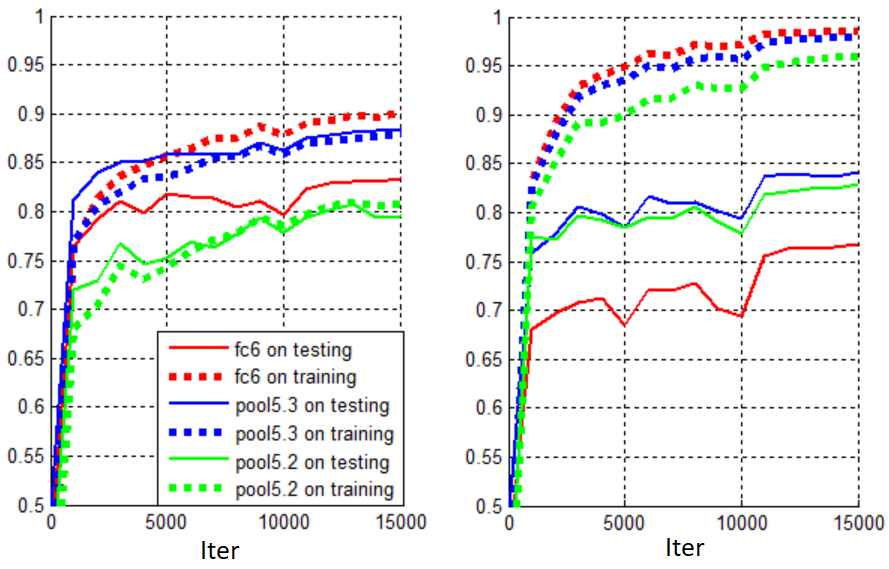}
  \caption{[Old version of Figure 3] R@1 performance of different layers on training and test set of Cars-196. Left: the variant trained with classification loss. Right: our baseline network trained with DML loss.}
  \label{fig:classification_old}
\end{figure}

\begin{figure*}[h!]
\begin{center}
  \includegraphics[width=0.9\linewidth]{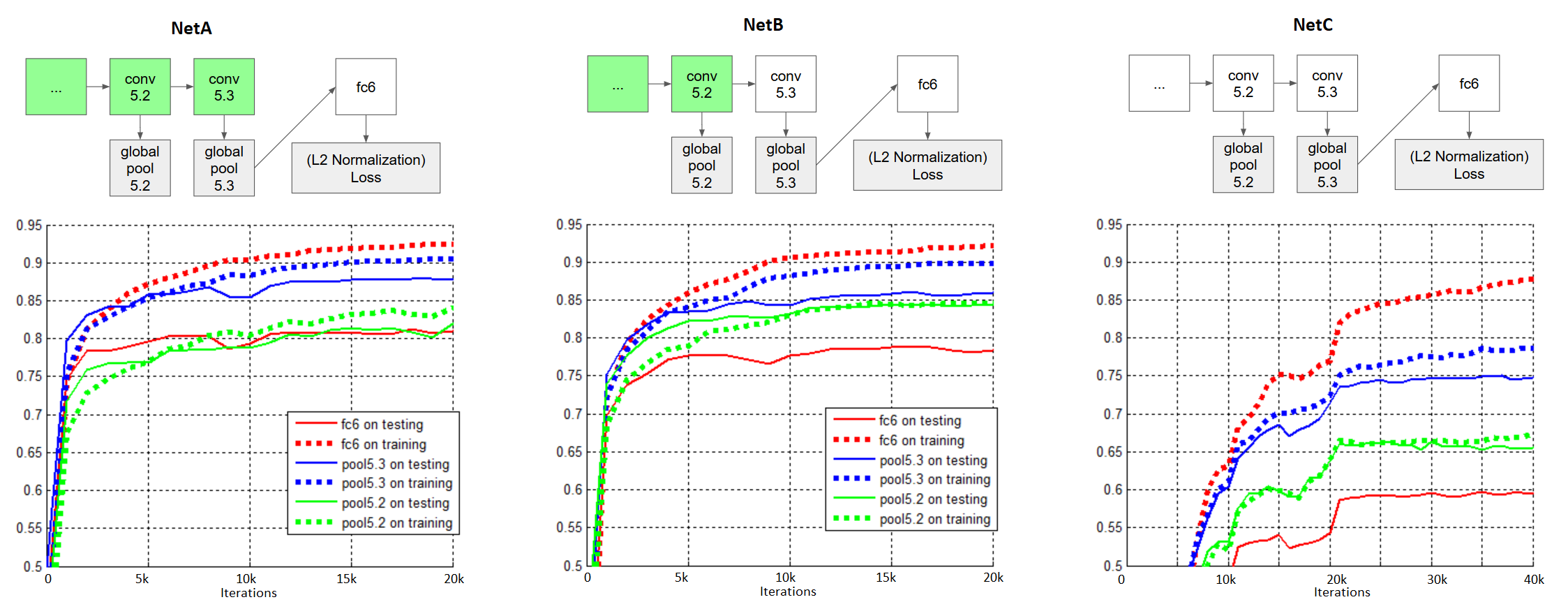}
  \vspace{5pt}
  \caption{[old version of Figure 4] R@1 performance of different layers on training and test set of Cars-196. Green box: pretrained layer, white box: initialized from scratch layer, gray box: parameterless layer. Best viewed in color.}
  \label{fig:analysize1_old}
\end{center}

\begin{center}
  \includegraphics[width=0.92\linewidth]{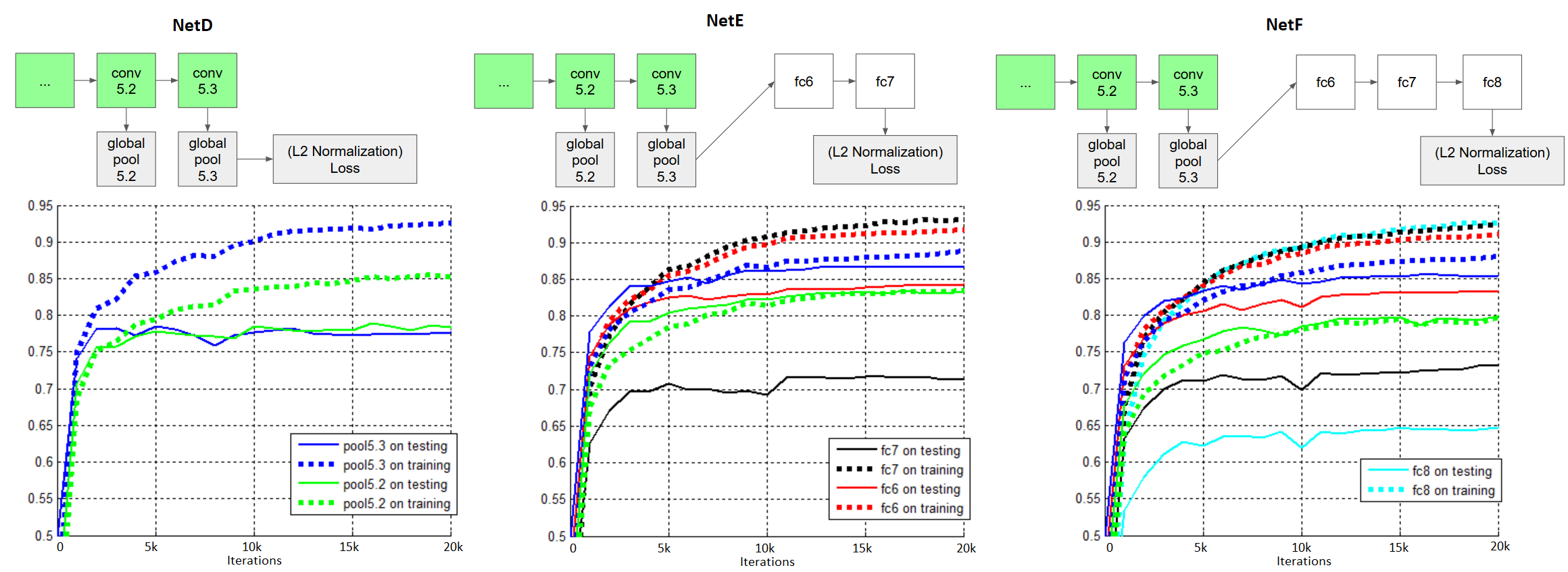}
  \vspace{5pt}
  \caption{[old version of Figure 5] R@1 performance of different layers on training and test set of Cars-196. Compared to NetA in Figure 4, the position of the loss function is changed. Best viewed in color.}
  \label{fig:analysize2_old}
\end{center}
\end{figure*}

\begin{figure*}[h]
\begin{tabular}{ll}
\includegraphics[width=0.45\linewidth]{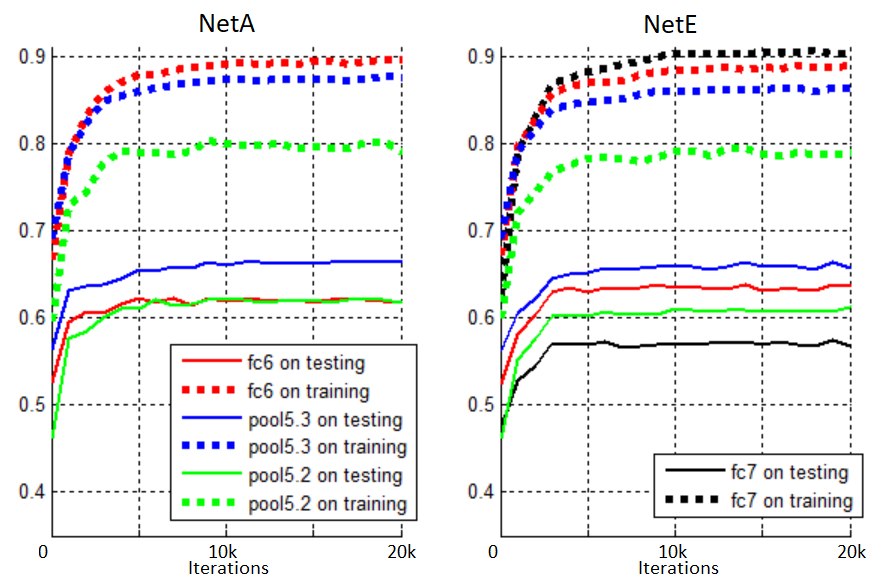}
&
\includegraphics[width=0.45\linewidth]{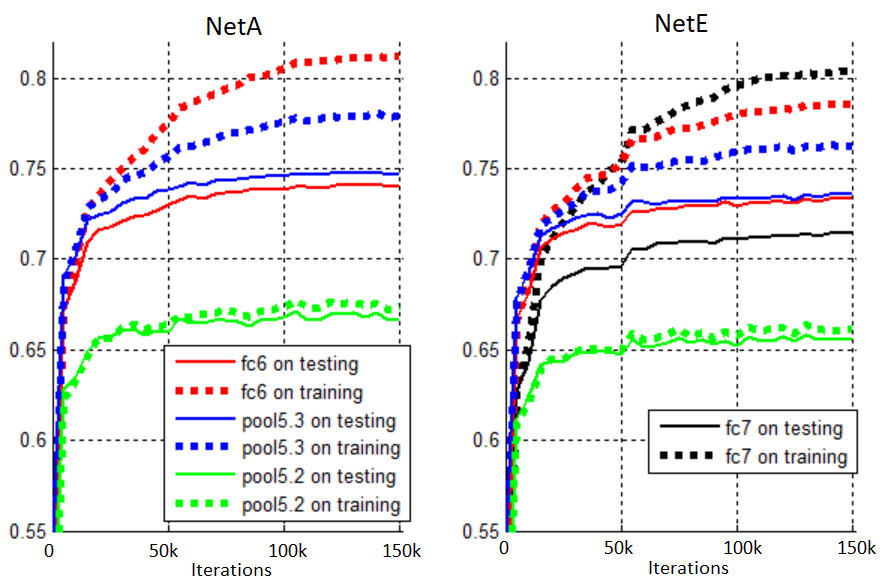}
\end{tabular}
\caption{[old version of Figure 8] R@1 performance of NetA and NetE on CUB-200-2011 (left) and Stanford Online Product (right)}
\label{fig:cub_old}
\end{figure*}

These are some old figures that are harder to read, but useful to use as reference when reproducing the result. Figure \ref{fig:classification_old}, \ref{fig:analysize1_old}, \ref{fig:analysize2_old}, \ref{fig:cub_old}.

Our code is publicly available at the website https://github.com/lugiavn/generalization-dml.

\section{Application to cross-view image retrieval}

We tried applying the finding here to the cross-view image retrieval task \cite{vo2016localizing}, but not successful. We speculate that this is because the search is cross different image domains. While they share the same high level image feature space, low level feature might be very different such that using layers before the final one results in incompatible features between street view images and overhead view images.

\newpage

\end{document}